 \documentclass[final,5p,times,twocolumn,authoryear]{elsarticle}

\usepackage{amssymb}
\usepackage{lipsum}
\usepackage{amsmath}
\journal{Neural Computing and Applications}

\begin{document}

\begin{frontmatter}

%% Title, authors and addresses

%% use the tnoteref command within \title for footnotes;
%% use the tnotetext command for theassociated footnote;
%% use the fnref command within \author or \affiliation for footnotes;
%% use the fntext command for theassociated footnote;
%% use the corref command within \author for corresponding author footnotes;
%% use the cortext command for theassociated footnote;
%% use the ead command for the email address,
%% and the form \ead[url] for the home page:
%% \title{Title\tnoteref{label1}}
%% \tnotetext[label1]{}
%% \author{Name\corref{cor1}\fnref{label2}}
%% \ead{email address}
%% \ead[url]{home page}
%% \fntext[label2]{}
%% \cortext[cor1]{}
%% \affiliation{organization={},
%%            addressline={}, 
%%            city={},
%%            postcode={}, 
%%            state={},
%%            country={}}
%% \fntext[label3]{}

\title{Semantic Feature Segmentation for Interpretable Predictive Maintenance in Complex Systems}

%% use optional labels to link authors explicitly to addresses:
%% \author[label1,label2]{}
%% \affiliation[label1]{organization={},
%%             addressline={},
%%             city={},
%%             postcode={},
%%             state={},
%%             country={}}
%%
%% \affiliation[label2]{organization={},
%%             addressline={},
%%             city={},
%%             postcode={},
%%             state={},
%%             country={}}

\author{Emilio Mastriani, Alessandro Costa, Federico Incardona, Kevin Munari, Sebastiano Spinello}
\affiliation
{organization={INAF, Osservatorio Astrofisico di Catania},
            addressline={Via S. Sofia 78}, 
            city={Catania},
            postcode={95123},
            country={Italy}}

\begin{abstract}
%% Text of abstract

Predictive maintenance in complex systems is often complicated by the heterogeneity and redundancy of monitored variables, which can obscure fault-relevant information and reduce model interpretability. This work proposes a semantic feature segmentation framework that decomposes the monitored feature space into a canonical component, expected to retain the dominant predictive information, and a residual component containing structurally peripheral signals. The segmentation is defined through domain-informed criteria and sets up monitoring variables into functional groups reflecting operational mechanisms such as throughput, latency, pressure, network activity, and structural state. To evaluate the effectiveness of this decomposition, we adopt a predictive perspective in which expected predictive risk is used as an operational proxy for task-relevant information. Experimental results obtained through time-aware cross-validation show that the canonical space consistently achieves lower predictive risk than the residual space across multiple temporal configurations, indicating that the semantic segmentation concentrates the most relevant information for fault anticipation. In addition, the canonical segments exhibit significantly stronger intra-segment coherence than inter-segment dependence, and this structural organization remains stable after redundancy reduction. When compared with the full feature space and with a Principal Component Analysis (PCA) representation, the canonical space carries out comparable predictive performance and furthermore preserves the semantic meaning of the original variables. These findings suggest that semantic feature segmentation provides an interpretable and information-preserving decomposition of monitoring signals, enabling competitive predictive performance without sacrificing the operational interpretability required in predictive maintenance applications.

\end{abstract}

%%Graphical abstract
%\begin{graphicalabstract}
%\includegraphics{grabs}
%\end{graphicalabstract}

%%Research highlights
%\begin{highlights}
%\item Research highlight 1
%\item Research highlight 2
%\end{highlights}

\begin{keyword}
%% keywords here, in the form: keyword \sep keyword, up to a maximum of 6 keywords
Predictive Maintenance \sep Semantic Feature Segmentation \sep Feature Space Decomposition\sep Distributed Systems Monitoring \sep Fault Prediction\sep  Imbalanced Data\sep Time-Aware Cross-Validation\sep Complex Systems

%% PACS codes here, in the form: \PACS code \sep code

%% MSC codes here, in the form: \MSC code \sep code
%% or \MSC[2008] code \sep code (2000 is the default)

\end{keyword}

\end{frontmatter}

%\tableofcontents

%% \linenumbers

%% main text

\section{Introduction}
\label{introduction}
The increasing availability of high-frequency monitoring data in industrial systems has significantly expanded the potential of predictive maintenance (PdM) strategies in recent years. Modern infrastructures, ranging from distributed computing systems to complex electromechanical platforms, are instrumented with many heterogeneous sensors and software-level metrics, producing high-dimensional time series that capture multiple aspects of system behavior (\cite{shi_modelling_2023}). While this abundance of data provides opportunities for early fault detection, it also introduces substantial challenges related to feature heterogeneity, redundancy, and background noise that may obscure signals relevant to failure dynamics (\cite{ndao_explainable_2026}).

A common assumption in various data-driven PdM approaches is that the predictive performance can be improved by leveraging as much information as possible, either through the direct use of all available features or via automated feature extraction techniques, such as dimensionality reduction and representation learning (\cite{LIANG2025127675}). However, in complex systems characterized by heterogeneous subsystems and partially overlapping dynamics, not all features contribute equally to the prediction tasks. In particular, a significant portion of the observed variables may encode background processes, redundant information, or weakly related signals that do not directly contribute to the fault anticipation. In our specific scenario - a containerized Cassandra database environment monitored through multiple JVM and system-level metrics - several features evolve according to distinct operational mechanisms (cumulative counters, latency percentiles, connection states, and memory pressure), and their indiscriminate use risks to obscure the signals truly indicative of the fault. The indiscriminate use of such features can lead to increased model complexity, reduced interpretability, and, occasionally, degraded predictive performance (\cite{jachymczyk_improved_2024}).

In this context, the identification and isolation of task-relevant information turns into a central issue. Classical approaches address this issue through feature selection or dimensionality reduction methods, aiming to maintain the most informative data components. Despite being effective in various settings, these techniques are typically driven by statistical criteria and often lack a clear connection to the underlying system structure (\cite{ROSSI2025113940}). Therefore, the selected features may be difficult to interpret and may not reflect the meaningful functional components of the system.

In this study, we propose an alternative perspective based on semantic feature segmentation, in which the feature space is partitioned into two complementary subspaces: a \textit{canonical} space designed to capture variables with high anticipatory capacity for fault events and a \textit{residual} space comprising the remaining features. Unlike purely statistical selection methods, this segmentation is manually designed and grounded in domain expertise, reflecting a functional taxonomy of the system monitoring metrics, such as cumulative throughput, latency, pressure, network state, and structural indicators, each associated with a specific operational mechanism. This expert-driven partition ensures that the segmentation is not merely \textit{statistical} but \textit{semantically} anchored in the system architecture, preserving interpretability and physical meaning.

The central hypothesis of this study is that this segmentation induces a functional decomposition of the system in favor of the prediction task. More specifically, we assume that the canonical space concentrates most of the task-relevant information, whereas the residual space contributes marginally or redundantly. To investigate this hypothesis, we adopted a predictive framework in which the expected risk, estimated via time-aware cross-validation and log-loss, was used as an operational measure of performance and, indirectly, of information relevance.

The analysis proceeds in two complementary directions. First, we compared the predictive performance of models trained separately on the canonical and residual spaces, showing that the canonical representation systematically achieves a lower risk, thus supporting its dominant role in fault prediction. Second, we examine the interaction between the two spaces and show that the residual space neither anticipates faults nor provides consistent conditional reinforcement in high-risk regimes identified by the canonical model. These results are further investigated through comparisons with the full feature space and a standard dimensionality reduction baseline based on Principal Component Analysis (PCA), allowing us to assess whether the proposed segmentation offers advantages over conventional feature engineering approaches.

Therefore, the contributions of this study are twofold. From a methodological perspective, we introduced a structured framework for evaluating feature relevance in complex systems based on semantic segmentation and predictive risk analysis. From an interpretative perspective, we provide empirical evidence supporting a hierarchical organization of the feature space, in which an information-rich core can be distinguished from a widely uninformative background. This dual contribution is particularly relevant in the PdM setting, where both predictive performance and interpretability play crucial roles in supporting operational decision-making (\cite{hakami_strategies_2024}).

Overall, the proposed approach offers a clear way to reduce complexity, improve interpretability, and maintain predictive effectiveness in high-dimensional industrial monitoring scenarios, contributing to a more structured understanding of the information flow in complex systems.

\section{Problem formulation}
Before presenting the formal mathematical framework, we briefly introduce the desired properties of the decomposition we aim to achieve. The \textit{segment separability} property requires that features belonging to the same semantic segment exhibit greater internal coherence than those belonging to different segments. The \textit{robustness of segment separability under redundancy reduction} requires that the separability condition persists even after the removal of highly collinear or redundant features from the canonical space. \textit{Functional decomposition of the system} states that semantic segmentation induces a decomposition of the system with respect to the prediction task, in which the canonical space captures more task-relevant information than the residual space. The fourth property, \textit{canonical space dominance and residual space marginality}, specifies that the residual space neither possesses significant predictive ability nor provides consistent conditional reinforcement in high-risk regimes identified by the canonical model. These four properties provide a conceptual foundation for the mathematical formulation introduced in the rest of this section.
\subsection{Segment separability}
Let 
\(F_{1}=\{f_{11}, f_{12},..., f_{1n}\}, F_{2}=\{f_{21}, f_{22},..., f_{2n}\}, ..., F_{m}=\{f_{m1}, f_{m2},..., f_{mn}\}\)
be the sets of values of the features \(f_1, f_2, ..., f_n\) along the temporal range \(t=1, 2, ..., m\). The set of values along the temporal axis can be written as:

\[F=\bigcup_{i=1}^{m}F_i=\bigcup_{\substack{j=1 \\ k=1}}^{\substack{j=m \\ k=n}}f_{jk}\]

Indicating with \(F_C\) the set of features that should contain strong signals of the complex system and with \(F_R\) the set of the remaining features, we have that \(F_C \cup F_R = F, F_C \cap F_R = \emptyset\). \(F_C\) is named the ‘\textit{canonical}' set, while \(F_R\) represents the set of ‘\textit{residual}’ features. At this point, we can partition the \(F_C\) set into a number of \(p\) segments \((S)\)containing the features \textit{semantically related} (\cite{xu_few-shot_2024}), that is the generic segment \(S_k\) is defined as \(S_k=\{(f_i,...,f_j) | sim(f_a,f_b)=1 \forall \ f_a,f_b \in S_k)\}\) where

\[sim(f_a,f_b)=\begin{cases} 1, & f_a \ and \ f_b \ are \ semantically \ related \\ 0, & f_a \ and \ f_b \ are \ NOT \ semantically \ related \end{cases}\]

As stated in the introduction, in this stage of the study, the semantic correlation is manually defined, according to the expert of field knowledge. 
By definition we get that \(F_C=\bigcup_{i=1}^{p}S_i, S_i \cap S_j=\emptyset, if  \ i \ne j\). 
Considering any two features belonging to \(F_C\) , let’s indicate with \(\rho_{sp}(f_i, f_j)\) the coefficient of the Spearman correlation between \(f_i\) and \(f_j\) (\cite{spearman_proof_1904}). The distribution of all the intra-segment correlations is defined by all the couple of features belonging to the \textit{same segment}:

\[\Omega_k^{in}=\{\rho_{sp}(f_i, f_j): f_i, f_j \in S_k, i \ne j \}\]

 The average correlation between all the features in \(S_k\) is represented by the Intra Canonical Correlation measure, and is calculated with the following equation:
 
\[ICC_k=\frac{2}{n_k(n_{k-1})}\sum_{f_i, f_j \in S_k}\rho_{sp}(f_i, f_j)\]

where \(n_k\) indicates the number of features in the \(k\) segment. The term 
\[ICC_{macro}=\frac{1}{p}\sum_{i=1}^{p}ICC_i\]
 represents the average value of all the averages ICC over all the segments. Representing with 
\[\Omega^{in}=\bigcup_{k=1}^{p}\Omega_{k}^{in}\]
 the set of all the distributions of correlations in \(F_C\) , the term 
\[ICC_{micro}=\frac{1}{|\Omega^{in}|}\sum_{k=1}^{p}ICC_k\]
 represents the average value of all the intra-segment correlations pooled. 

Let’s consider any two segments \(S_a, S_b \in F_C\)  with \(a \ne b\) . We represent the set of all the correlations between \textit{different segments} with 
\[\Omega_{ab}^{out}=\{\rho_{sp}(f_i,f_j) | f_i \in S_a, f_j \in S_b, i \ne j\} \]

The average value of inter-segment correlation between any two segments \(S_a, S_b \in F_C\)  with \(a \ne b\) is given by:

\[ICOR_{ab}=\frac{1}{n_a n_b}\sum_{f_i \in S_a}\sum_{f_j \in S_b}\rho_{sp}(f_i, f_j)\]

In the same way  we proceeded with ICC, the ICOR can be calculated in two ways too:

\[ICOR_{macro}=\frac{1}{\binom{p}{2}}\sum_{a < b}ICOR_{ab}\] representing the average value of all the couple of segments, and 

\[ICOR_{micro}=\frac{1}{|\Omega^{out}|}\sum_{a < b} \sum_{f_i \in S_a} \sum_{f_j \in S_b} \rho_{sp}(f_i, f_j)\]
representing the average of all the inter-segment correlation pooled with , where 
\[\Omega^{out}=\bigcup_{a<b}\Omega_{ab}^{out}\]

Remembering that in our context the ICC value represents the internal coherence of every subsystem, while the ICOR value quantify how much the subsystems stay separated from each other, to prove the “segment separability” means to verify with a non-parametric test of Mann-Whitney U. (\cite{mann_test_1947}, \cite{wilcoxon_individual_1945}) the following relation:

\[H_0: ICC \le ICOR \ vs \ H_1: ICC > ICOR\]

That is, features belonging to the same semantic segment show greater significant internal coherence than relationships between different segments, proving that semantic segments are truly separate and coherent, capturing distinct subsystems of the system.

\subsection{\textbf{Robustness of segment separability under redundancy reduction}}
In the previous section, we defined the canonical space  \(F_C\) as a limited set, structured as a collection of semantic segments  \(\{S_{k=1}^{p}\}\) such that each segment represents one coherent sub-family of features with respect to a specific dimension of the system (latency, state, pressure, etc.). Formally, the segments constitute a partition of the canonical space of features:

\[\bigcup_{i=1}^{p}S_i=F_C\]

Based on this structure, we introduced the concept of separability using two functional operators derived from the correlations between the features: Intra-Canonical Correlation (ICC), which assesses the coherence within the segments, and Inter-Canonical Correlation (ICOR), which quantifies the dependency between different segments. The distinction between correlation (within-segment coherence) and redundancy (across-segment duplication) is central to feature selection theory, where high correlation among features may indicate either meaningful shared dynamics or superfluous duplication \cite{sun_correlation-redundancy_2024}.
Let’s now introduce the operator of \textit{redundancy reduction} 
\[P: F_C \to \tilde{F}_{C}\]
where  \(\tilde{F}_{C} \subseteq F_{C}\) represents the space of features after the \textit{pruning} task. Such an operator can be intended as a \textit{selective projection} that eliminates the variables being redundant or highly collinear, and preserves the variables retained informative with regards to the threshold of correlation intra-segment. The removal of redundant features based on correlation analysis is a well-established practice to mitigate multiple collinearity and improve model interpretability while preserving predictive information (\cite{capraz_feature_2024}). 
One crucial point while applying the redundancy reduction is that the segmentation does not need to be defined again, it is simply induced to the new reduced space. For each original segment  \(S_k\) we get that
\[\tilde{S}_k = S_k \cap \tilde{F}_C\]
which represents the restriction of the segment to the features “\textit{persisted}” through elimination of redundancies. This step ensures the semantic coherence of the model: the structure of the segments stays unaltered, while the cardinality just changes and so the internal information density changes too. This property is particularly desirable in explainable feature selection frameworks, where preserving the semantic interpretability of selected features is a primary goal (\cite{LIANG2025127675}).
The correlation metrics have to be recalculated in the reduced space without any formal modification, defining them with

\begin{equation}
    \label{eq:ICC_P>ICOR_P}
    ICC^{(P)} > ICOR^{(P)}
\end{equation} 
as the post-pruning counterparts of the original measures, obtained performing the same correlation operation (Spearman) to the new couples of features that belong respectively to  \(\tilde{S}_k\) and to \(\tilde{S}_k \times \tilde{S}_l\) with \ \(k \ne l\). Demonstrating that the separability is stable with respect to the transformation induced by P, is equivalent not only to verify  that the inequality (\ref{eq:ICC_P>ICOR_P}) persists, but also to observe that the relation of \textit{quasi-invariance} in the comparison with the original situation still continues to be valid

\[ICC^{(P)} \approx ICC \ and \ ICOR^{(P)} \approx ICOR\]
This result would imply that the observed structure is not an artifact due to the presence of duplications or spurious dependencies, but an intrinsic property of the semantic organization of the feature space, highlighting the stability of the property. In fact, the segmentation would result not only valid, but also robust with respect to the transformation that eliminates redundant information.

\subsection{Functional decomposition of the system}
Once segment separability and the robustness of separability under redundancy reduction have been established, the next step is to demonstrate that this organized representation of the system has an operational meaning in the predictive maintenance context. Specifically, we aim to show that the introduced semantic decomposition leads to a reduction in predictive risk with respect to the residual space.
Let \(X \in \mathbb{R}^{d}\) denote the space of the observed features, and \(Y \in \{0,1\}\) as the fault variable. Let \(X = (X_C, X_R)\) be the decomposition induced by the semantic segmentation, where \(X_C\) is the canonical space, and \(X_R\) the residual space. We introduce two models \(f_C\) and \(f_R\), belonging to the same functional class, and trained respectively on \(X_C\) and \(X_R\). The corresponding expected risks are defined as:

\[\mathcal{R}_C = \mathbb{E}[\mathcal{L}(f_C(X_C), Y)] \ and \ \mathcal{R}_R = \mathbb{E}[\mathcal{L}(f_R(X_R), Y)] \]
where \(\mathcal{L}\) is an appropriate loss function (e.g., log-loss) indicating how far the prediction is from the true value, while the expectation  \(\mathbb{E}\) denotes the average over the data distribution. Hence,  \(\mathcal{R}_C\) and \(\mathcal{R}_R\) represent the average prediction error of the models trained on the canonical and residual space, respectively.
Let  \(I(X;Y)\) denote the mutual information between X and Y, which quantifies of how much knowing X reduces uncertainty about Y: the lower the uncertainty, the higher  \(I(X;Y)\) will be. Our objective is to establish that observing  \(X_C\) reduces uncertainty about Y more than  \(X_R\) does, i.e. :

\begin{equation}
    I(X_C; Y) > I(X_R; Y)
    \label{eq:I(X_C; Y) > I(X_R; Y)}
\end{equation}
Direct estimation of mutual information is, however, unreliable in high-dimensional and low-fault regimes. Therefore, we adopt a predictive perspective: under sufficiently expressive models, a lower achievable predictive risk is indicative of a higher amount of task-relevant information. In this sense, an empirical observation of
\[\mathcal{R}_C < \mathcal{R}_R\]
provides evidence that the canonical space contains more fault-relevant information than the residual space, and is therefore consistent with the inequality (\ref{eq:I(X_C; Y) > I(X_R; Y)}). Once this property is verified, semantic segmentation is not merely a descriptive tool, but induces a functional decomposition of the system with respect to the prediction task, separating a structured, information-rich component from a less informative background.

\subsection{Canonical space dominance and residual space marginality}
Let us give a complex system whose features are partitioned into two distinct subspaces: the canonical space \(X_C\) , containing the features with high anticipatory predictive capacity for fault events, and the space \(X_R\), consisting of the remaining, less directly informative features. Let indicate with the equation (\ref{eq:p_C-AND-p_R}) the conditional predictive probabilities on the two spaces, where \(Y(t+\Delta)\) indicates the appearance of a fault at the future step \(\Delta\).
\begin{equation}
    \begin{aligned}
        & p_C(t+\Delta) = \Pr\bigl(Y(t+\Delta)=1 \mid X_C(t)\bigr), \\
        & p_R(t+\Delta) = \Pr\bigl(Y(t+\Delta)=1 \mid X_R(t)\bigr)
    \end{aligned}
    \label{eq:p_C-AND-p_R}
\end{equation}
We wish to show that, for any time horizon \(\Delta > 0\), the residual space does not possess significant anticipatory predictive ability compared to the canonical space. Formally, it proposes to test that the following properties reported in (\ref{AUC05}) are valid, excluding a relevant predictive capacity of the residual space.

\begin{equation}
    \begin{aligned}
        & \ R(X_R) \gtrsim R_{baseline}, \\
        & AUC(X_R) \approx 0.5
    \end{aligned}
    \label{AUC05}
\end{equation}

Furthermore, we would to prove that the conditional covariance between \(p_R\) and \(p_C\) in high risk cases is small and not consistently positive across the models and horizons, demonstrating that does not systematically reinforce high-risk predictions. Formally, it proposes to verify the property (\ref{Cov}) where \(\theta_C\) represents a high risk threshold predicted by canonical space.  
\begin{equation}
    \forall \Delta >0, \ Cov(p_R(t+\Delta), p_C(t+\Delta) | p_C(t+\Delta) > \theta_C) \lesssim 0
    \label{Cov}
\end{equation}
These results \textbf{highlight the primacy of the canonical space}: the residual space adds little independent information, confirming that predictive strategies should rely primarily on \(X_C\).

\section{Methods}
\subsection{Pipeline overview}
The implemented pipeline performs structured preprocessing and semantic transformation of multivariate time-series monitoring data collected from a Cassandra node, with the purpose of organizing raw telemetry metrics into an interpretable canonical feature space, validating the coherence of the semantic segmentation, and subsequently reducing redundancy through intra-segment pruning.

\begin{figure}
    \centering
    \includegraphics[width=0.5\linewidth]{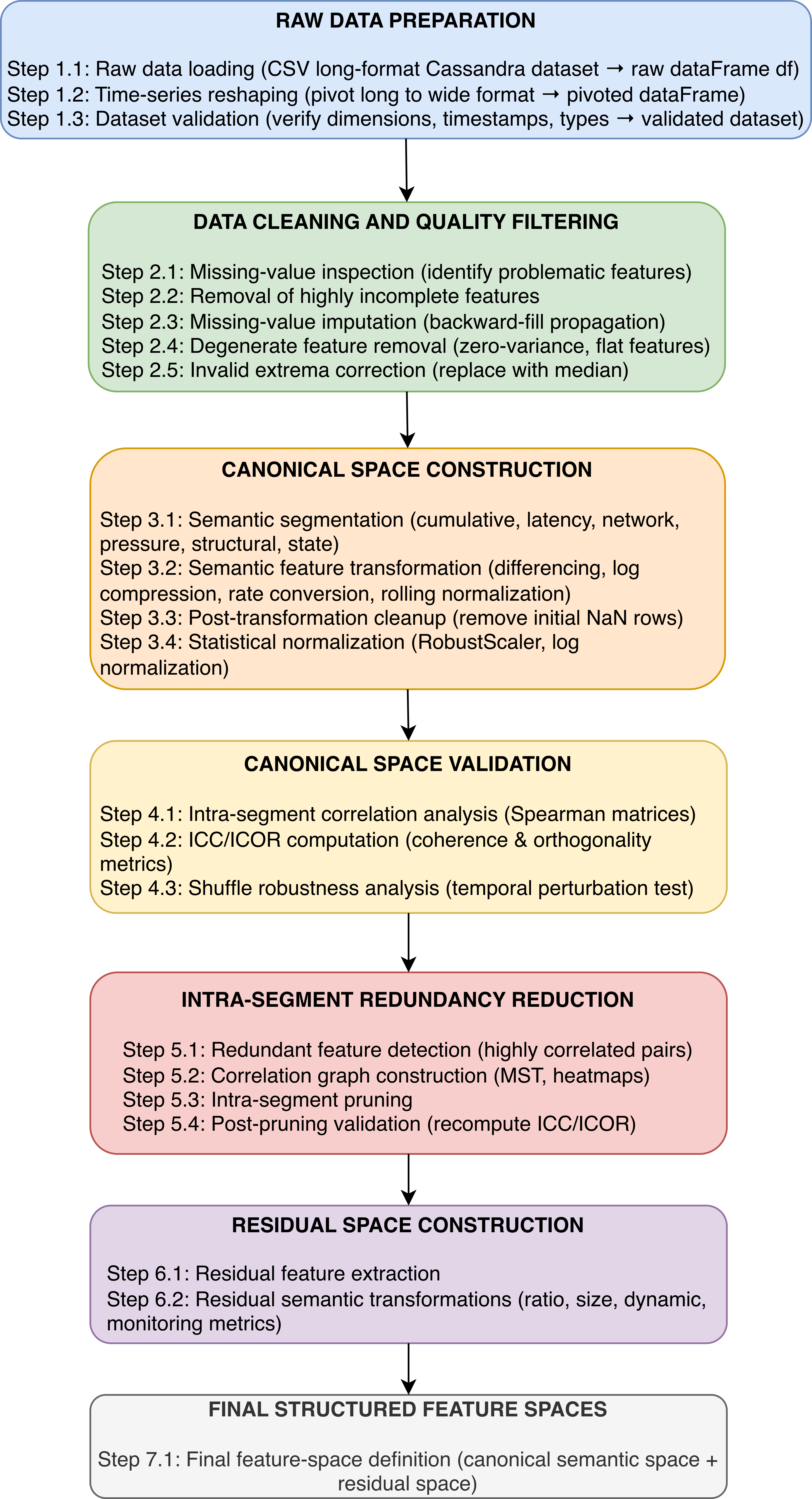}
    \caption{Schematic workflow of the data processing pipeline for handling Cassandra monitoring data. The pipeline comprises seven main stages: raw data preparation, data cleaning and quality filtering, construction and validation of the semantic canonical space, intra-segment redundancy reduction, residual space construction, and final definition of structured feature spaces.}
    \label{fig:flowchart}
\end{figure}

The workflow (Figure \ref{fig:flowchart}) begins with the ingestion of raw metrics and reshaping of the dataset into a time-aligned wide representation. After a preliminary cleaning phase covering missing-value treatment, removal of degenerate variables, and correction of invalid extrema, the pipeline applies a semantic transformation strategy, remodeling the metrics according to their operational meaning (e.g., counters, latency indicators, pressure metrics, structural indicators, and state variables).

Once transformed, the resulting canonical space is evaluated through intra-segment and inter-segment correlation analyses, allowing the computation of global structural indicators (ICC and ICOR) that evaluate the internal coherence and external orthogonality of semantic segmentation. Based on these correlations, a graph-based pruning stage removes redundant intra-segment variables while preserving the semantic structure of the feature space.

Finally, the pruned canonical representation was used as the basis for constructing a residual feature space, where additional transformations were applied to the remaining metrics according to their statistical and operational behavior. This establishes the complete feature organization required for the subsequent functional decomposition and predictive analysis. Figure  \ref{fig:flowchart} reports the full pipeline's steps. 

\subsection{Experimental Test Environment}
According to our experience, one of the frequent fault state in complex systems is related to the file descriptor exhaustion and resource contention. Due to this reason, our stress test pushed to gain this kind of fault. 
We conducted the experimental validation on a containerized Apache Cassandra cluster deployed on Linux-based hosts. During the test session, six instances of opcuasimulatormon [RIF] produced normal background traffic, simulating 988 monitoring points. The total monitoring duration was 6032 min. Fault injection was performed using the Distributed Infrastructure Stress and Fault Injector (DISFI) tool (\cite{mastriani_emilio_2026_18627402}), a custom traffic generator designed to stress Cassandra connection management and internal thread pools. The DISFI tool supports connection storms, read storms, intentional connection leaks, and CPU spikes, enabling the reproducible triggering of file descriptor exhaustion and request timeouts. All injected faults were time-stamped and logged, allowing for precise alignment with the observed system anomalies. In order to achieve the target anomaly usually observed in production, we designed three distinct fault scenarios to trigger the file descriptor exhaustion. The \textbf{first} scenario generates a high rate of short-lived client connections, forcing each thread to repeatedly open and close Cassandra sessions, producing irregular connection bursts that exhaust the server-side connection pools and file descriptors. The \textbf{second} scenario increases both the operation rate per thread and the logical sensor space, creating overlapping write bursts with frequent reconnections, saturating memtables, and commit log synchronization. The \textbf{third} scenario combines aggressive write traffic, forced reconnections, and concurrent read queries over a very large sensor space (100,000 sensors), explicitly targeting the read-write interference under unstable connection behavior. Each test was performed multiple times over the total monitoring period. We ran the DISFI tool eight times in total: we performed tests 1, 2, and 3 twice at regular intervals of about 12 h, setting the duration parameter to 4 min for each execution. Between fault injections, normal traffic sessions lasted 8–16 h, producing a \textit{naturally unbalanced} dataset in which fault events represented a small fraction of the overall temporal window. The following confirmed faults were observed: on 2025-12-12 at 00:09:22 (12 threads in timeout), on 2025-12-13 at 12:22:02 (12 threads), on 2025-12-14 at 00:26:13 (17 threads), at 00:26:55 (15 threads), at 00:27:54 (4 threads), at 00:28:12 (11 threads), at 00:28:46 (4 threads), and on 2025-12-15 at 00:33:41 (2 threads). Container resource limits were configured with soft and hard limits of 1,048,576 open files, and a global limit from the kernel of 6,301,104. Cassandra thread pools were configured with concurrent\_reads, concurrent\_writes, and concurrent\_counter\_writes set to 32, and concurrent\_compactors set to 1. The JVM maximum open files limit was 1,048,576. For a complete description of the testbed configuration, fault injection parameters, and detailed test execution logs, refer to the supplementary material (Appendix A).
\subsection{Data preparation}
The raw data extracted from InfluxDB exhibited a heterogeneous tabular structure, with separate tables generated for each Java Virtual Machine metric and an additional table containing Apache Cassandra metrics, all of which were concatenated in a single export. To convert this raw structure into a format suitable for the framework, the preprocessing phase first involved the unification of the exported tables through an ad-hoc script, followed by a pivot transformation from a long to a wide format, as detailed in Appendix A. This process produced an intermediate dataset of 2,157,331 rows and 17 columns, which was reshaped into a data-frame of 12,120 rows and 137 columns, indexed by the timestamp.
After confirming the numerical consistency of all the variables, the dataset was inspected for missing values. Features characterized by an almost complete absence of observations were discarded, whereas columns containing only a few missing entries were completed using backward-fill propagation. Additional filtering was applied to eliminate variables with null variance and features affected by a high proportion of missing values. These preprocessing steps reduced the dataset dimensionality from 137 to 83 variables while preserving the temporal structure required for the subsequent stages.  
\subsection{Rationale for Semantic Transformation}
The monitored software infrastructure can be regarded as a complex system composed of multiple interacting components, represented by heterogeneous metrics describing processes, file descriptors, memory usage, network activity, and internal service states. In such systems, the observed behavior emerges from nonlinear interactions among variables over time, and faults rarely depend on a single metric; rather, they arise from collective dynamics involving multiple resources and processes (\cite{bousdekis_proactive_2017}).

Under these conditions, data preprocessing cannot be considered a purely statistical operation. Since the monitored variables reflect distinct operational mechanisms, applying a uniform transformation to the entire feature space may distort their original meaning and alter the relationships that characterize the system dynamics. In particular, global normalization may introduce interference among heterogeneous variables, reducing interpretability and weakening the consistency of the representation.

To preserve the operational significance of the monitored signals, the preprocessing strategy adopted in this work introduces a \textit{semantic transformation stage} before statistical normalization. This transformation reorganizes the feature space according to the functional role of the variables, ensuring that subsequent normalization is performed only within semantically homogeneous groups. This approach preserves the interpretability of the signals while reducing distortions caused by the coexistence of heterogeneous metrics in the same statistical space.

\subsection{Canonical Space Definition}
To achieve a semantically coherent representation of the monitored infrastructure, the original feature space was partitioned into groups of variables sharing similar operational behavior. The grouping criterion was based on the functional meaning of the metrics and on the type of dynamics they represent within the system.

Following this principle, the monitored variables were organized into semantic segments corresponding to cumulative processes, latency indicators, pressure and backlog dynamics, network connectivity states, system state variables, and structural descriptors of the software infrastructure. Variables not satisfying the coherence criteria required for these semantic groups were preserved in a residual information space.

This segmentation defines the canonical space adopted in the proposed framework. Each segment represents a semantically homogeneous subspace in which the variables share comparable dynamics and operational interpretation. The residual space keeps the remaining features for ex-post validation and exploratory analysis, without affecting the internal coherence of the primary semantic segments.

\subsection{Transformations within Semantic Segment}
Once the canonical space has been defined, statistical transformations are applied independently within each semantic segment. This choice reflects the fact that the monitored variables are not generic statistical observations but operational signals whose distributions are shaped by the underlying mechanisms of the infrastructure (\cite{cai_data-driven_2017}).

Classical global transformations aimed at reducing skewness or enforcing distributional symmetry may alter the operational meaning of these signals, since asymmetries, accumulations, and abrupt variations often represent meaningful characteristics of the system rather than statistical anomalies. Applying transformations within semantically homogeneous segments preserves the local structure of the data while enabling comparability among variables that share the same functional behavior.

The transformation process therefore maintains the semantic consistency of the canonical space and avoids introducing distortions between variables generated by different operational mechanisms. On this basis, each semantic segment is processed through a dedicated transformation strategy consistent with the dynamics of the variables it contains.

\subsection{Segment-Specific Transformation Strategies}
Following the semantic partitioning of the canonical space, each segment was processed through a dedicated transformation pipeline designed to preserve the operational meaning of the corresponding variables while ensuring numerical comparability. This segment-wise transformation framework follows recent predictive maintenance preprocessing approaches that emphasize the preservation of the physical meaning of monitored signals while improving numerical robustness during model preparation (\cite{zhao_deep_2019}, \cite{carvalho_systematic_2019}). The complete set of transformations adopted for each semantic group is summarized in Table \ref{tab:semantic_transformations}, where each segment is associated with a logical transformation, semantic transformation operator, and corresponding normalization strategy.

\begin{table*}[t!]
\centering
\footnotesize
\setlength{\tabcolsep}{3pt}
\renewcommand{\arraystretch}{1.25}
\caption{Semantic transformation and normalization strategies adopted for each feature segment in the canonical space.}
\label{tab:semantic_transformations}

\begin{tabular}{p{1.5cm} p{7.0cm} p{2.7cm} p{2.5cm} p{1.8cm}}
\hline
\textbf{Segment} & \textbf{Features} & \textbf{Logical Transformation} & \textbf{Semantic Transformation} & \textbf{Normalization} \\
\hline

Cumulative &
\raggedright
processcpusecondstotal, jvmclassesloadedtotal,\\
jvmclassesunloadedtotal, jvmthreadsstartedtotal,\\
cassandranetworkreceivebytes, cassandranetworktransmitbytes,\\
cassandraCommitLogCompletedTasks, cassandraCompactionCompletedTasks,\\
cassandrareadlatencycount, cassandrawritelatencycount,\\
jvmgccollectionsecondscount, jvmgccollectionsecondssum &
\raggedright Monotonic Counter Rate (MCR) &
$\max(0, x_t-x_{t-1})$ &
Robust Scaler \\

Latency &
\raggedright
cassandrareadlatency99th, cassandrawritelatency99th,\\
cassandrareadtotallatency, cassandrawritetotallatency &
\raggedright Logarithmic Tail Compression (LTC) &
$\log(1+x)$ &
Robust Scaler \\

Pressure &
\raggedright
cassandraColumnFamilyPendingCompactions,\\
cassandraTablePendingCompactions,\\
cassandraColumnFamilyPendingFlushes,\\
cassandraTablePendingFlushes,\\
cassandraCompactionPendingTasks &
\raggedright Baseline Stress Ratio (BSR) &
$\dfrac{x_t}{\tilde{x}_t^{(w)}+\epsilon}$ &
$\log(1+x)$ \\

Network &
\raggedright
cassandraconnstateestab, cassandraconnstatetimewait,\\
cassandraopensockets &
\raggedright First-order difference &
$\dfrac{\max(0,x_t-x_{t-1})}{\Delta t}$ &
No Scaling \\

State &
\raggedright
processresidentmemorybytes, processvirtualmemorybytes,\\
jvmmemorybytesused, jvmmemorybytescommitted,\\
jvmmemorypoolbytesused, jvmmemorypoolbytescommitted,\\
jvmbufferpoolusedbytes, jvmbufferpoolcapacitybytes,\\
cassandraColumnFamilyLiveDiskSpaceUsed,\\
cassandraTableLiveDiskSpaceUsed,\\
cassandraColumnFamilyTotalDiskSpaceUsed,\\
cassandraTableTotalDiskSpaceUsed &
\raggedright Global Baseline Deviation (GBD) &
$\dfrac{x_t-\tilde{x}}{x}$ &
No Scaling \\

Structural &
\raggedright
cassandraColumnFamilyLiveSSTableCount,\\
cassandraTableLiveSSTableCount,\\
cassandraColumnFamilyMemtableLiveDataSize,\\
cassandraTableMemtableLiveDataSize &
\raggedright Rolling Baseline Drift (RBDR) &
$\dfrac{x_t-\tilde{x}_t^{(w)}}{\tilde{x}_t^{(w)}+\epsilon}$ &
No Scaling \\

\hline
\end{tabular}
\end{table*}

Cumulative metrics were transformed through the Monotonic Counter Rate (MCR), which converts monotonically increasing counters into interval-based activity rates by computing the non-negative first difference. This transformation removes the dependence on the cumulative magnitude and preserves local variations that reflect changes in system activity. Because the resulting distributions remained highly asymmetric and burst-driven, robust scaling was subsequently applied to ensure comparability across variables while preserving the relative intensity of activity spikes.

Latency metrics were processed using Logarithmic Tail Compression (LTC), which applies logarithmic mapping to compress the heavy-tailed distributions commonly observed in latency-related measurements. This transformation reduces the scale heterogeneity while preserving the relative structure of the latency peaks. To further improve the comparability among variables without introducing sensitivity to extreme values, robust scaling was applied after semantic transformation.

Pressure metrics were transformed using the Baseline Stress Ratio (BSR), which expresses each observation relative to a rolling operational baseline. This transformation emphasizes the deviations associated with congestion or backlog accumulation, preserving the stress-related dynamics relevant for predictive maintenance. Because these variables are characterized by sparse peaks and strong asymmetry, logarithmic compression was subsequently applied to stabilize the numerical range without suppressing stress events.

Network metrics were converted into activity rates using a first-order differencing operator with non-negative clipping, thereby transforming cumulative connection counters into interpretable event-rate variables. This approach preserves meaningful variations while handling counter resets and transient negative jumps. Owing to the bounded and sparse nature of these transformed variables, no additional statistical scaling was applied.

State metrics, including memory and storage occupancy indicators, were transformed using the Global Baseline Deviation (GBD), which expresses each value as a relative deviation from its baseline operating condition. This transformation removes the scale dependency while preserving the interpretation of the variables as deviations from the nominal resource utilization state. Because the transformed values were already dimensionless and directly comparable, no additional normalization was required.

Structural metrics were processed using the Rolling Baseline Drift Ratio (RBDR), which measures the relative deviation from a rolling structural baseline. This transformation highlights gradual structural drift while preserving the temporal dynamics of the storage subsystem. Because the transformed values were already expressed as normalized relative deviations, no further statistical scaling was applied.

The statistical rationale underlying these segment-specific choices, including the treatment of skewness, sparsity, and scale heterogeneity, is presented in Appendix A. 

\subsection{Intra-segment Correlation Analysis and Redundancy Reduction}
After the semantic transformation stage, each segment of the canonical space contains variables describing a homogeneous operational aspect of the monitored infrastructure. Although this segmentation reduces semantic heterogeneity, redundant information may still be present among variables representing the same underlying process at different aggregation levels or through closely related measurements. To identify these dependencies and reduce feature duplication, an intra-segment correlation analysis was performed independently for each semantic segment. For each transformed segment, a correlation matrix was computed using Spearman’s rank correlation coefficient. This choice was adopted uniformly across all segments to ensure methodological consistency and provide robustness against non-Gaussian distributions and nonlinear monotonic dependencies, which are common in infrastructure monitoring data. Restricting the analysis to semantically homogeneous groups ensures that the resulting dependencies preserve operational meaning and avoid spurious associations between heterogeneous variables. To achieve a compact representation of the internal dependency structure, the absolute correlation matrix of each segment was converted into a distance matrix and used to construct a Minimum Spanning Tree (MST). This representation preserves the minimal set of strongest links connecting the variables within each semantic segment, enabling the identification of redundant metrics while maintaining the primary operational relationships. Based on the MST structure, redundant variables were identified as those exhibiting near-equivalent behavior regarding other features that describe the same operational phenomenon (\cite{li_feature_2018}, \cite{mantegna_hierarchical_1999}). In such cases, a single representative variable was maintained according to its operational interpretability and ability to preserve the informational content of the subsystem. This pruning strategy was applied independently within each semantic segment to preserve the semantic organization of the canonical space while reducing multicollinearity and dimensional redundancy. Redundancy analysis revealed that duplicated information was mainly associated with metrics describing the same process at multiple aggregation levels, including Cassandra table- and column-family level statistics, mirrored network counters, and overlapping JVM resource indicators. Removing these redundant variables yielded a reduced semantic representation that preserved the principal operational dimensions of the monitored infrastructure while improving the compactness and stability of the feature space. Table \ref{tab:feature_pruning_summary} summarizes the variables removed during the intra-segment redundancy reduction process and the representative signals kept for each semantic segment. Detailed MST structures and segment-wise pruning decisions are reported in Appendix A.

\begin{table}[t!]
\caption{Summary of intra-segment redundancy reduction after MST-based correlation analysis.}
\label{tab:feature_pruning_summary}
\centering
\scriptsize
\setlength{\tabcolsep}{3pt}
\renewcommand{\arraystretch}{1.15}

\begin{tabular}{p{1.2cm} p{2.4cm} p{3.0cm}}
\hline
\textbf{Segment} & \textbf{Removed Features} & \textbf{Retained Operational Signals} \\
\hline

Pressure &
Column-family pending compaction/flush metrics &
Compaction backlog, flush backlog, scheduling pressure \\

Cumulative &
GC count, outbound network traffic &
CPU load, incoming traffic, JVM activity, commit log workload \\

Network &
Open sockets &
Established connections, transient socket events \\

Latency &
Read total latency &
Tail latency indicators, cumulative write latency \\

State &
Redundant JVM/disk usage metrics &
Memory usage, buffer occupancy, storage allocation \\

Structural &
Column-family SSTable/memtable metrics &
SSTable organization, memtable occupancy \\

\hline
\end{tabular}
\end{table}

\subsection{Residual Feature Space}

The residual feature space contains variables not meeting the semantic coherence criteria for canonical segments. These features may still contain useful information, but they do not exhibit sufficiently stable operational semantics to justify integration into the main semantic structure. Rather than forcing them into the canonical representation, they are preserved as an auxiliary information space dedicated to validation, robustness analysis, and the identification of behaviors not explicitly modeled within the primary semantic segments (\cite{tzionis_review_2025}).

To preserve this exploratory role, the residual features are not subjected to the same semantic transformation framework adopted for the canonical space. Instead, a lightweight preprocessing strategy is applied to improve numerical stability while avoiding the introduction of artificial structure. This ensures that the residual space remains methodologically independent from the semantically organized feature space and can serve as an unbiased reference layer for post validation (\cite{data9050069}).

Although residual features are excluded from the main semantic segmentation, they exhibit recurrent statistical patterns that allow a lightweight organization based on empirical behavior rather than operational meaning. For this reason, the residual space was partitioned into four weakly structured families: bounded ratio metrics, size and volume metrics, weak dynamic or quasi-static metrics, and monitoring or auxiliary metrics. This \textit{soft taxonomy} supports the selection of coherent preprocessing operations without imposing semantic assumptions on variables that do not belong to the canonical operational structure.

Bounded ratio metrics include stable variables with naturally constrained domains, such as Bloom filter indicators and compression ratios. Size and volume metrics comprise memory- and byte-related quantities characterized by scale heterogeneity and positive skewness. Weak dynamic metrics correspond to variables exhibiting limited temporal variability, such as JVM thread states and slowly varying counters. Monitoring and auxiliary metrics include technical indicators related to the monitoring infrastructure rather than to the operational state of the system itself.

The preprocessing strategies associated with these residual families are summarized in Table \ref{tab:residual_taxonomy}. The detailed operational treatment applied to each residual family is reported in Appendix A. This organization preserves the exploratory nature of the residual feature space while ensuring numerical consistency and methodological transparency.
\begin{table}[t!]
\caption{Lightweight taxonomy and preprocessing strategies adopted for the residual feature space.}
\label{tab:residual_taxonomy}
\centering
\scriptsize
\setlength{\tabcolsep}{2pt}
\renewcommand{\arraystretch}{1.15}

\begin{tabular}{p{1.6cm} p{2.2cm} p{2.0cm} p{1.6cm}}
\hline
\textbf{Residual family} & \textbf{Feature type} & \textbf{Transformation} & \textbf{Normalization} \\
\hline

Ratio \& bounded &
Bounded ratios and stable indicators &
None or $\sqrt{x}$ &
Conditional Z-score \\

Size \& volume &
Memory and byte-related variables &
$\log(1+x)$ &
Robust Scaler \\

Weak dynamic &
Low-dynamic counters and states &
None or first difference &
Minimal scaling \\

Monitoring &
Technical support metrics &
None &
None \\

\hline
\end{tabular}
\end{table}
% Required package:
% \usepackage{makecell}

\section{Results}
\subsection{Segment Separability: semantic segmentation is an emergent property of coordinated temporal dynamics}

Results emerging from the analysis and reported in Table \ref{tab:segment_separability} show a clear separation between intra- and inter-segment correlations in canonical space. Particularly clear is the gap between these two measures: the mean intra-segment correlation, denoted as \(ICC_{micro}\), reaches a value of 0.3820, while the inter-segment correlation, \(ICOR_{micro}\), stands at only 0.0791. The difference, approximately 0.30, is substantial and not marginal, suggesting the presence of a structured internal organization. The statistical test further confirms the significance of this result, with a p-value close to zero, suggesting that the greater internal coherence of the segments is not because of chance.

\begin{table}[ht]
\centering
\small
\setlength{\tabcolsep}{5pt}
\caption{Summary of global and segment-level correlation statistics in the canonical feature space.}
\label{tab:segment_separability}
\begin{tabular}{lcc}
\hline
Statistic & Value & Notes \\
\hline
$ICC_{micro}$ & 0.3820 & Mean intra-segment correlation \\
$ICOR_{micro}$ & 0.0791 & Mean inter-segment correlation \\
$\Delta$ & 0.3029 & $ICC_{micro} - ICOR_{micro}$ \\
$p$-value & $<0.001$ & Mann--Whitney U test \\
State ICC & 0.6750 & Highest intra-segment coherence \\
Latency--State ICOR & 0.4246 & High mean, low median \\
\hline
\end{tabular}
\end{table}

At the structural level, differentiated characteristics emerge among the segments. State segment stands out for its extremely high internal coherence, with a value of approximately 0.67 and several near-perfect correlations, suggesting strong internal redundancy but also a notably solid structure. The other segments — Cumulative, Latency, Pressure, and Structural — show lower mean correlation values, while maintaining systematically higher coherence compared to that observed between different segments. The latter, with a mean ICOR that remains low (approximately 0.07), supports the interpretation that the identified segments capture distinct coordinated behaviors rather than arbitrary groupings of variables.
An aspect of particular interest concerns the relationship between the Latency and State segments. Here, the mean inter-segment correlation is relatively high (approximately 0.42), but the median remains at very low values. This discrepancy suggests that there are a few strong but not systematic links, resulting in a typical example of outlier structure rather than a global dependence between the two segments. This pattern does not undermine the overall separation of the segments, but enriches the understanding of their point-wise interactions.
In summary, the results show a clear separation between intra- and inter-segment correlations. The canonical segments exhibit significantly higher internal coherence (ICC) compared to relationships between segments (ICOR), with a considerable effect size (\(\Delta \approx 0.30\)). This supports the interpretation that the proposed semantic segmentation captures structurally coherent feature groups, and not merely arbitrary aggregations of features. Detailed intra- and inter-segment correlation statistics for each canonical segment and segment pair are reported in Appendix A.
\begin{figure}
    \centering
    \includegraphics[width=0.75\linewidth]{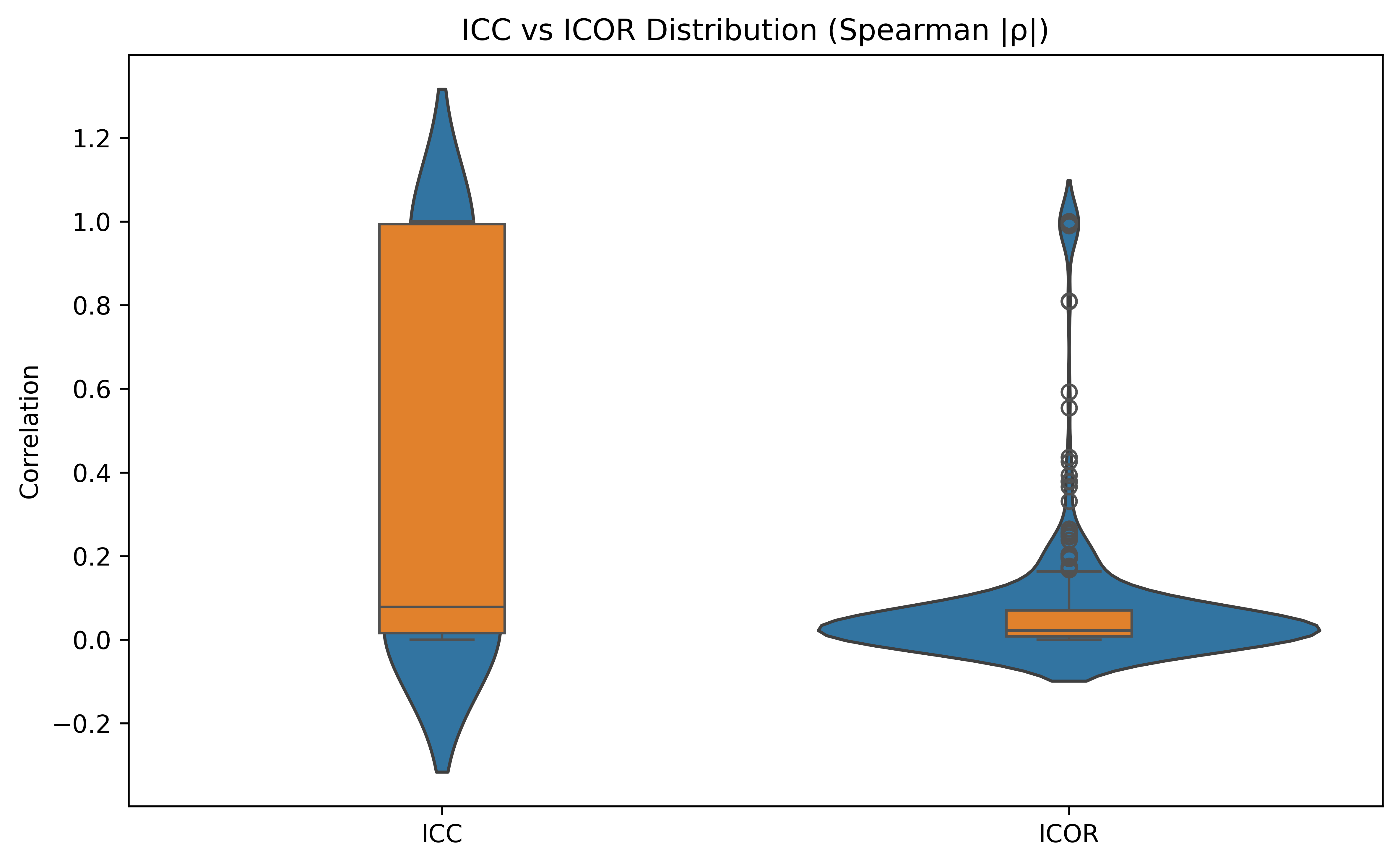}
    \caption{Violin plot comparing the distributions of intra-segment correlation (ICC) and inter-segment correlation (ICOR) in the canonical feature space. The ICC distribution is shifted toward higher values and exhibits a long upper tail, indicating the presence of strongly coordinated feature pairs within semantic segments. In contrast, the ICOR distribution is concentrated near zero with only a few outliers, suggesting generally weak correlations between features belonging to different segments. This distributional separation provides visual evidence of the stronger internal coherence of canonical segments compared to cross-segment relationships.}
    \label{fig:violin_plot}
\end{figure}
In addition, the violin plot reported in Figure \ref{fig:violin_plot} shows that ICC is characterized by a distribution shifted toward higher values with a long tail (because of very strong correlations), while ICOR appears concentrated near zero with few outliers. 

\begin{figure}
    \centering
    \includegraphics[width=0.75\linewidth]{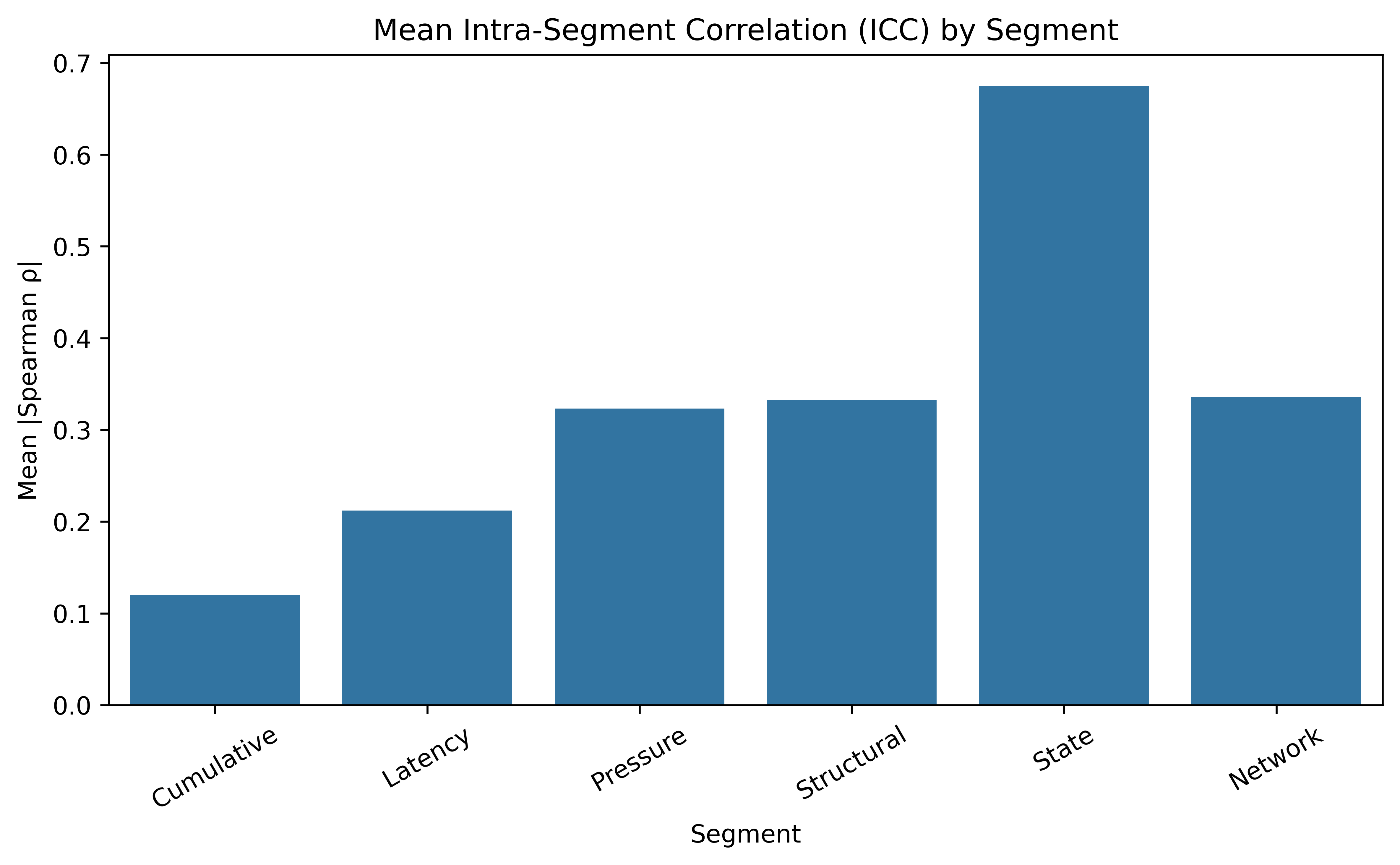}
    \caption{Mean intra-segment correlation values for each canonical segment, compared with the global mean inter-segment correlation (ICOR). The State segment exhibits the highest internal correlation, indicating a strong degree of redundancy but also a highly coherent internal structure. The remaining segments show lower but still consistently positive intra-segment correlation values, all remaining above the global ICOR baseline. This pattern indicates that canonical segments differ in their degree of internal cohesion, yet each preserves a stronger internal dependency structure than that observed across segments, supporting the validity of the semantic segmentation while motivating redundancy reduction in highly correlated segments.}
    \label{fig:bar_plot}
\end{figure}

The bar chart reported in Figure \ref{fig:bar_plot} shows how the State segment dominates, highlighting strong redundancy (resolved with subsequent pruning) and presenting the other segments with a weaker but real structure. The graph highlights that not all segments are equal, but all are above ICOR.
While these results provide strong evidence of structural separability, they do not yet clarify whether this organization arises from static statistical properties or from underlying temporal dynamics. To further validate that segment separability arises from coherent temporal dynamics, we performed a \textit{temporal perturbation test} based on feature-wise circular shifts (\cite{yuan_rigorous_2024}). This transformation preserves univariate statistical properties and the autocorrelation structure of each individual time series while selectively destroying cross-feature temporal alignment. Under this perturbation, the original relationship between intra- and inter-segment correlations is no longer preserved. In particular, the results show a clear \textit{inversion} with \(ICC_{shift}=0.061\) and \(ICOR_{shift}=0.0343\).

This behavior is highly informative: once temporal synchrony across features is disrupted, correlations between features belonging to different segments become artificially dominant, while within-segment coherence collapses. This remarks that the observed separability is not a trivial statistical artifact, but depends on the underlying temporal organization of the system, supporting the conclusion that the identified segment structure is driven by coherent temporal dynamics. Importantly, this inversion does not contradict the original findings; rather, it strengthens them. If segment separability were merely a consequence of static statistical structure, it would persist under temporal perturbations. Its disappearance under temporal misalignment proves that the segmentation captures genuine dynamical organization. 
This result provides strong evidence that segment separability is intrinsically linked to the temporal coordination of system variables and cannot be attributed solely to temporal autocorrelation or shared trends.

\subsection{Robust segment separability under redundancy reduction}

The comparison between the pre- and post-pruning phases yields a very clear result that is particularly significant from a scientific standpoint. In the original configuration, the canonical space exhibits a sharp separation between intra-segment and inter-segment correlations: mean ICC values (\(\approx0.38\)) at the micro level) are significantly higher than ICOR values   (\(\approx0.08\)), with a substantial difference  (\(\approx0.30\)) that is strongly supported by statistical testing. This confirms that semantic segmentation is not arbitrary, but reflects a genuine structured organization of features, in which variables belonging to the same segment share a much more pronounced common dynamic compared to variables belonging to different segments.

\begin{table}[htbp]
\centering
\caption{Comparison of global intra-segment correlation (ICC), inter-segment correlation (ICOR), and their difference before and after redundancy pruning. Although intra-segment correlation decreases after pruning, the ICC--ICOR gap remains substantial and statistically significant, indicating that semantic segment separability is preserved after redundancy reduction.}
\label{tab:pruning_separability}
\begin{tabular}{lcc}
\hline
\textbf{Metric} & \textbf{Pre-pruning} & \textbf{Post-pruning} \\
\hline
ICC$_{micro}$ & 0.3820 & 0.3108 \\
ICOR$_{micro}$ & 0.0791 & 0.0745 \\
$\Delta = ICC - ICOR$ & 0.3029 & 0.2363 \\
p-value & $<0.001$ & $<0.001$ \\
\hline
\end{tabular}
\end{table}

Following redundancy pruning, there is a quantitative but not qualitative change in the picture. As expected, Table \ref{tab:pruning_separability} shows a reduction in mean ICC values  (\(\approx0.31\)) and, to a much lesser extent, in ICOR values  (\(\approx0.07\)). The decrease in ICC is a natural consequence of removing highly correlated features, which contributed to increasing the measured internal coherence of segments. However, what is crucial is that the separation between the two distributions remains well-defined: the ICC–ICOR difference remains substantial  (\(\approx0.24\)) and statistically significant. Pruning reduces the intensity of internal correlation but does not alter the structural hierarchy between intra- and inter-segment correlations.

A peculiar aspect emerges when observing the \textit{stability} of ICOR. Inter-segment correlations remain substantially unchanged, suggesting that pruning acts predominantly within segments, eliminating local redundancies without introducing mixing between different semantic structures. This is consistent with the idea that segments represent distinct informational subspaces, and that their separation is not an artifact because of the presence of duplicate or highly collinear features.

Concurrently, some segments — particularly State — retain high internal coherence, maintaining very high levels of intra-segment correlation even after pruning. This behavior suggests that, in such cases, coherence does not depend on redundancy but on a genuine latent structure shared among features. Conversely, segments with few remaining elements after pruning exhibit a more pronounced reduction in ICC, stating that part of their cohesion was indeed supported by internal redundancy.

Overall, the fundamental result is that the property of segment separability is not only confirmed but proves robust regarding dimensionality reduction and redundancy removal. The structure observed in the canonical space does not collapse when features are compressed, but is preserved in a more parsimonious form.

\subsection{Functional Decomposition of the Complex System}

\begin{figure}
    \centering
    \includegraphics[width=0.75\linewidth]{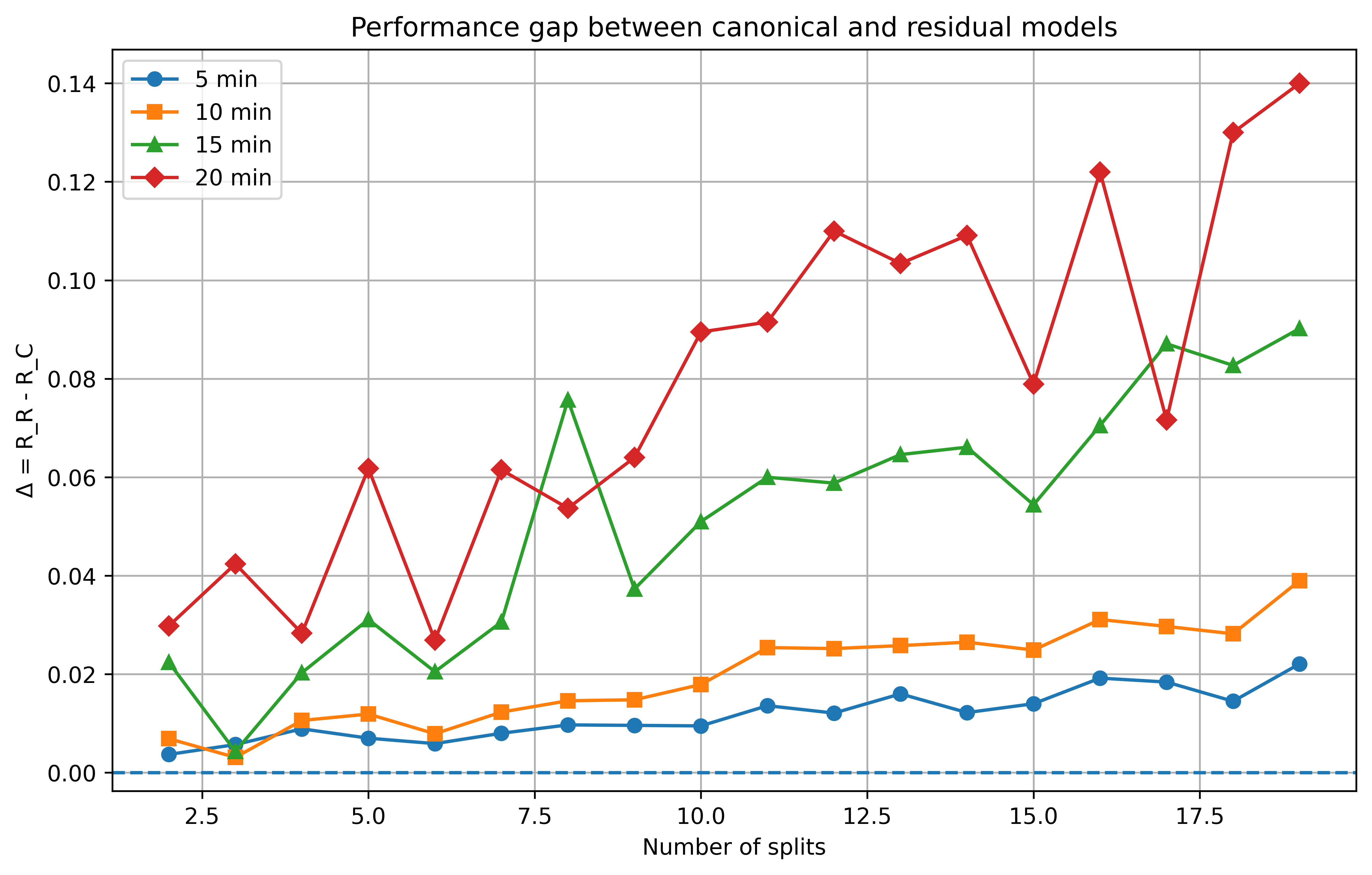}
    \caption{Performance gap between residual and canonical feature spaces across temporal configurations. The plot reports the difference $\Delta = R_R - R_C$ as a function of the number of temporal splits for multiple aggregation horizons (5, 10, 15, and 20 minutes). The gap remains consistently positive across all configurations, indicating that the canonical space systematically achieves lower predictive risk than the residual space. The magnitude of the gap increases with longer aggregation windows, suggesting that the advantage of the canonical representation becomes more pronounced at coarser temporal scales.}
    \label{fig:performance_gap}
\end{figure}

The empirical analysis conducted across multiple temporal configurations consistently supports the central hypothesis of this study that the predictive risk associated with the residual space is systematically higher than that got in the semantically structured canonical space. Across all experimental conditions, regardless of the number of temporal splits or aggregation window, the relation \(R_R>R_C\) was consistently observed. This inequality does not emerge as an isolated or contingent phenomenon but appears robust across the entire range of tested configurations, suggesting that there is a structural advantage linked to the canonical representation. This result is notably relevant when interpreted in the light of the heterogeneous nature of the features involved. The analyzed system integrates variables originating from different technological subsystems, each of which is characterized by distinct dynamics. In this context, semantic segmentation enables the isolation of a subset of variables that preserves the most task-relevant information while reducing the influence of background or redundant signals. The observed reduction in predictive risk within the canonical space can, therefore, be interpreted as evidence that this representation is not only more compact but also more aligned with the underlying structure of the system. The Figure \ref{fig:performance_gap} provides visual support for this result, showing that the gap \(\Delta=R_R - R_C\) remains consistently positive for all configurations considered.

\subsection{Comparative Evaluation of Semantic Segmentation Against Standard Feature Representations}

We assessed whether the proposed semantic segmentation method provides a meaningful advantage over standard feature engineering strategies by conducting a systematic comparative analysis across multiple representations of the feature space. Principal Component Analysis (PCA) is a widely adopted unsupervised dimensionality reduction technique that identifies orthogonal directions that maximize the variance in the data without explicitly accounting for task relevance ( \cite{jolliffe2002principal, shlens_tutorial_2014}). Besides the canonical \(X_C\) and residual \(X_R\) subspaces, we introduced two baselines commonly adopted in machine learning practice: the full feature space \(X_{Full}\) and a dimensionality-reduced representation obtained via PCA. All representations were evaluated under the same experimental protocol using a time-aware cross-validation scheme and a consistent predictive model, with performance measured in terms of the expected log-loss risk.

The results, aggregated across varying numbers of temporal splits, revealed a coherent and stable pattern. The canonical space consistently achieved a lower risk than the residual space, confirming that semantic decomposition effectively concentrates predictive information. Simultaneously, the full feature space does not consistently improve the predictive performance over the canonical representation, suggesting that the additional features introduce redundancy rather than a useful signal.

More critically, the comparison with the PCA baseline provides a direct answer to concerns regarding standard feature engineering approaches. The PCA-based representation yields a predictive performance that is consistently comparable to that of the canonical space. The differences between \(R_C\) and \(R_{PCA}\) are small across all splits and are not statistically significant, suggesting that both representations hold a similar amount of predictive information.

This comparison is relevant because PCA provides optimal linear compression in terms of variance preservation, whereas the proposed semantic segmentation is not variance-driven but \textit{structurally grounded}. The comparable performance between the two strategies suggests that task-relevant information is largely aligned with the semantically defined structure rather than purely with directions of maximal variance.

\begin{table}[htbp]
\centering
\caption{Average predictive risk (log-loss, lower is better) across different feature space representations. The canonical space achieves lower risk than the residual space, while exhibiting performance comparable to both the full feature space and the PCA-based representation. This indicates that semantic segmentation preserves the dominant predictive information while providing a more structured and interpretable decomposition.}
\label{tab:risk_comparison}
\begin{tabular}{lc}
\hline
\textbf{Representation} & \textbf{Mean Risk} \\
\hline
Canonical ($R_C$) & 0.155 \\
Residual ($R_R$) & 0.202 \\
Full ($R_{FULL}$) & 0.200 \\
PCA ($R_{PCA}$) & 0.163 \\
\hline
\end{tabular}
\end{table}

To facilitate interpretation, Table \ref{tab:risk_comparison} reports the average risk values across all considered splits (from 2 to 9), highlighting the relative positions of each representation.

The table confirms three key observations. First, the canonical space achieved a substantially lower risk than the residual space, reinforcing the interpretation of the latter as a background or noise-dominated component. Second, the full feature space did not provide a measurable advantage over the canonical representation, indicating that the segmentation did not discard relevant predictive information. Third, the PCA baseline performed similarly to the canonical space, suggesting that the proposed semantic decomposition is competitive with the optimal linear compression of the data.

From a statistical standpoint, none of the pairwise comparisons between the canonical, full, and PCA representations yielded significant differences under paired tests across the folds. This absence of significance should not be interpreted as a weakness but rather as an indication of robustness: the canonical representation maintains predictive performance across different temporal configurations without exhibiting instability or overfitting.

Overall, these findings support the conclusion that semantic segmentation constitutes an information-preserving and structurally meaningful transformation of the feature space. Although it does not lead to a measurable improvement in predictive accuracy over standard approaches, such as PCA, it achieves equivalent performance while introducing a structured and interpretable decomposition. In the context of predictive maintenance, where understanding the origin and role of signals is often as important as the predictive performance itself, this property represents a non-trivial advantage.
\subsection{Canonical Space Dominance and Residual Space Marginality }

The empirical evaluation was conducted across multiple prediction horizons \(\Delta\) and using three distinct model classes (Rolling LGBM, XGBoost, Random Forest) to ensure robustness with respect to the learning algorithm (\cite{LightGBM_Ke2017}, \cite{chen_xgboost:_2016}, \cite{breiman_random_2001}). The analysis focuses on the comparative predictive capacity of the canonical space \(X_C\) and residual space \(X_R\), as well as their potential interaction in high-risk regimes. 

\begin{figure}
    \centering
    \includegraphics[width=1\linewidth]{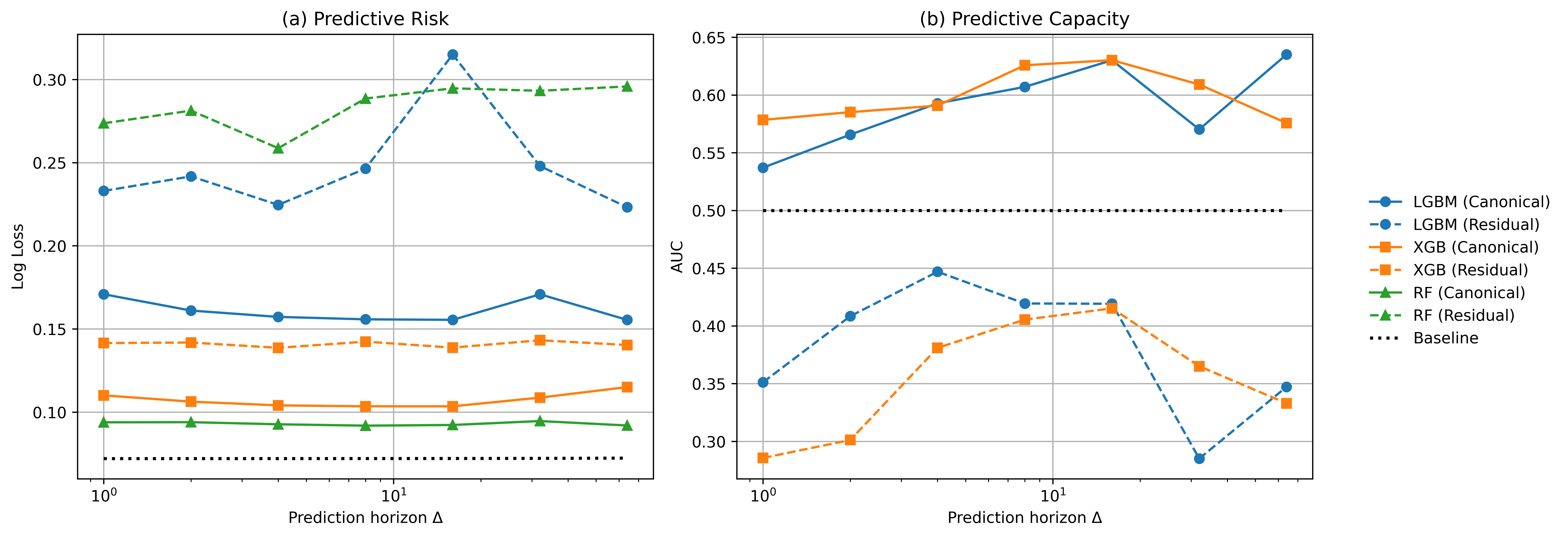}
    \caption{Predictive performance across prediction horizons \(\Delta\) for canonical \(X_C\) and residual \(X_R\) feature spaces using different models (Rolling LGBM, XGBoost, Random Forest).
(a) Predictive risk (log-loss): the canonical space consistently achieves lower risk than the residual space across all horizons, while the residual space remains close to or above the non-informative baseline.
(b) Discriminative performance (AUC): the canonical space maintains stable performance above baseline, whereas the residual space remains close to 0.5, indicating the absence of meaningful discriminative power }
    \label{fig:predictiverisk_capacity}
\end{figure}

Figure \ref{fig:predictiverisk_capacity} shows the predictive risk (log loss) and discriminative performance (AUC) as functions of the prediction horizon. A consistent pattern emerged across all models: the canonical space systematically outperformed the residual space for every value of \(\Delta\). In particular, \(R(X_C)\) remains consistently lower than \(R(X_R)\), while the AUC associated with \(R(X_C)\) remains above the baseline levels (typically in the range 0.55--0.63 depending on the model and horizon). In contrast, the residual space exhibited substantially higher predictive risk, often approaching or exceeding the non-informative baseline, and an AUC persistently close to 0.5. This behavior is stable across models and horizons, providing strong empirical support for condition expressed in (\ref{AUC05}), namely, that the residual space lacks meaningful anticipatory predictive capacity. 

\begin{figure}
    \centering
    \includegraphics[width=0.75\linewidth]{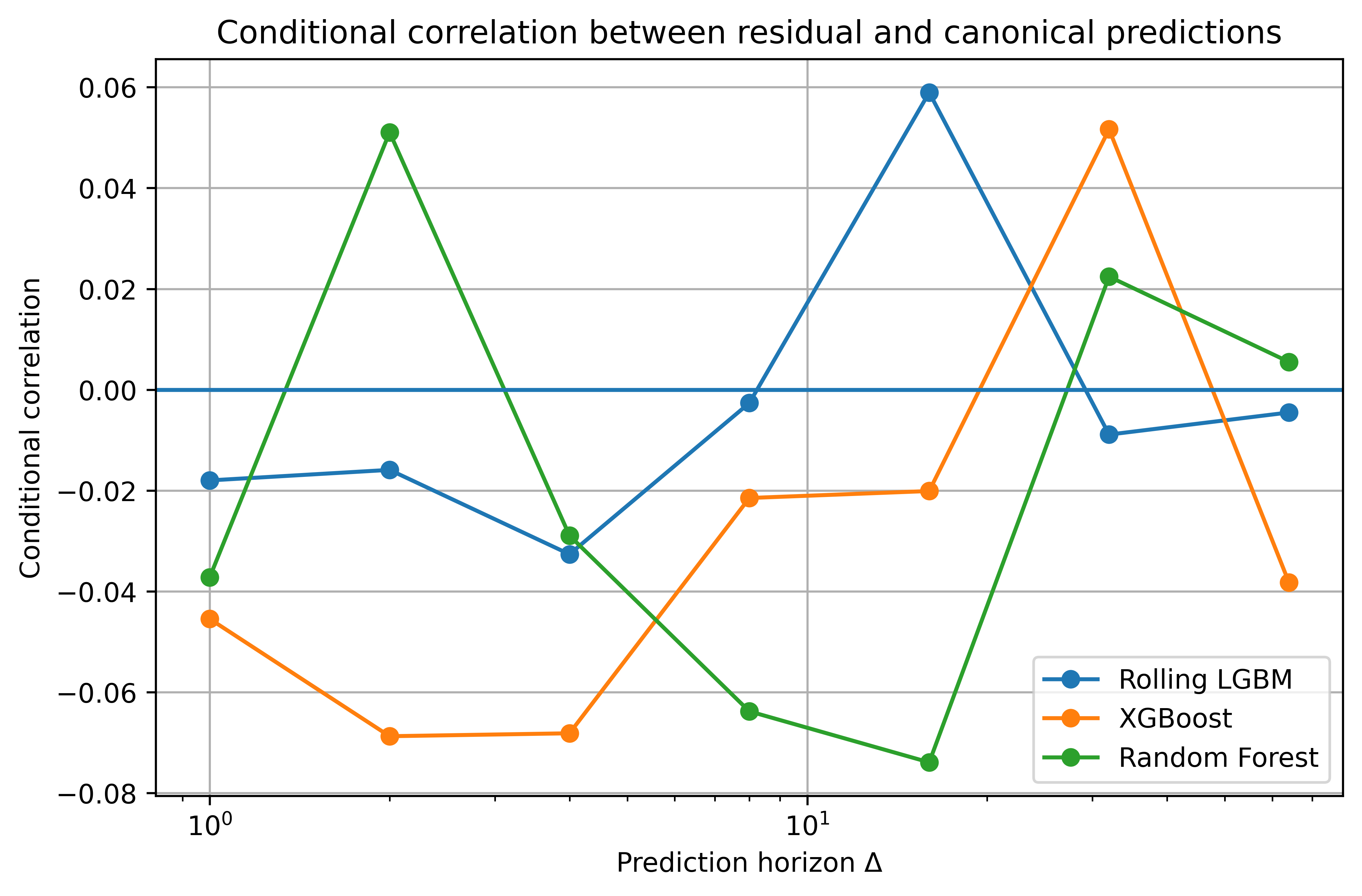}
    \caption{Conditional correlation between residual and canonical predictive outputs, \(corr(p_R, p_C \mid p_C > \theta_C)\), as a function of the prediction horizon \(\Delta\) across different models. The horizontal line at zero indicates the absence of linear dependence. Observed values fluctuate around zero without a stable or consistent pattern, indicating the absence of systematic co-movement between the two spaces in high-risk regimes. This suggests that the residual space does not provide reliable complementary information when the canonical model identifies elevated fault probability.}
    \label{fig:conditionalcorrelation}
\end{figure}

Beyond marginal performance, Figure \ref{fig:conditionalcorrelation} investigates the conditional relationship between the two spaces by evaluating the correlation \(corr(p_R, p_C \mid p_C > \theta_C)\) , that is, restricted to high-risk regimes, as identified by the canonical model. The results show that the conditional correlation fluctuates around zero, with a small magnitude and no consistent sign across models or horizons. Occasional positive values appear, but they are neither stable nor reproducible across the methods. This indicates the absence of a systematic co-movement between \(p_R\) and \(p_C\) under critical conditions. Consequently, condition presented in (\ref{Cov}) is empirically supported in its negative form, suggesting that the residual space does not provide reliable confirmation signals when the canonical space identifies a high fault probability. Taken together, these findings establish a clear dominance of the canonical space, both in terms of standalone predictive performance and in the absence of meaningful complementary information from the residual space. The residual features neither anticipated faults nor consistently reinforced high-risk predictions. From a predictive maintenance perspective, this result has a direct operational implication: modeling efforts and feature engineering can be primarily focused on the canonical space, as the residual space does not provide additional actionable information under the current partitioning scheme.

\section{Discussion}

The results reached in this study provide convergent empirical evidence that semantic feature segmentation captures a meaningful structural organization of the monitored system and that this organization has direct relevance for predictive maintenance. The canonical segmentation demonstrates three consistent characteristics throughout the various stages of analysis: inherent structural coherence, resilience when redundancy is reduced, and the maintenance of predictive information. Taken together, these results support the interpretation of semantic segmentation as an effective functional decomposition of the monitored feature space.

The primary important finding relates to the structural separability of the canonical segments. The empirical observation that intra-segment correlations are systematically higher than inter-segment correlations demonstrates that features grouped within the same semantic segment share a stronger coordinated behavior than features belonging to different segments. This effect remains statistically significant at the global level and is observed consistently across all canonical segments, although with different magnitudes. Particularly informative is the persistence of this pattern after temporal perturbation: when the temporal alignment among variables is intentionally disrupted, the relationship between intra- and inter-segment correlation collapses. This points out that the observed segment coherence is not merely a consequence of static distributional similarity, but rather emerges from coordinated temporal dynamics. In methodological terms, this result validates semantic segmentation as a representation of functional subsystems rather than a simple taxonomy of variables.

The redundancy-reduction analysis further strengthens this interpretation. After removing highly correlated features within each segment, the overall magnitude of intra-segment correlation decreases, as expected, but the structural separation between intra- and inter-segment dependencies remains clearly preserved. This finding is especially significant as it confirms that segment separability is not just a side effect of redundant or overly similar data. Instead, the persistence of the separability gap after pruning suggests that each segment retains an intrinsic internal organization even when redundant information is removed. In this sense, the canonical segmentation identifies subsystems characterized by genuine coordinated dynamics rather than by superficial collinearity.

The predictive experiments provide the second major line of evidence. Across all temporal validation configurations, the canonical space consistently achieves lower predictive risk than the residual space. This result suggests that semantic segmentation effectively concentrates the predictive information relevant to fault anticipation within the canonical subspace. The advantage of the canonical representation over the residual space is observed independently of the specific temporal split and remains stable across the tested predictive models, suggesting that it reflects a structural property of the representation rather than a model-dependent effect. From the perspective of predictive maintenance, this finding implies that the variables organized in the semantic segments capture the principal operational mechanisms associated with the onset of anomalous conditions.

Equally informative is the behavior of the residual space. Although the residual variables are not entirely devoid of information, their predictive performance remains consistently inferior to that of the canonical space and close to the non-informative baseline. In addition, the conditional dependence analysis shows that the predictions generated from the residual space do not exhibit a stable positive relationship with those generated from the canonical space in high-risk conditions. This lack of conditional co-movement shows that the residual features do not provide reliable complementary signals when the system approaches failure. Therefore, the residual space should not be interpreted as a meaningful auxiliary predictor under the current decomposition, but as a collection of weakly informative or structurally peripheral signals. This distinction is crucial: the residual space is not necessarily irrelevant in an absolute sense, but it does not contribute significant predictive value beyond the canonical representation.

Comparing with the full feature space and PCA-based dimensionality reduction offers more insight into the canonical representation’s nature. The fact that the predictive performance of the canonical space remains comparable to that of both the full representation and the PCA baseline emphasizes that semantic segmentation preserves most of the predictive information contained in the original feature space. This is a relevant result because it confirms that the proposed decomposition does not sacrifice predictive capacity while introducing a structured and interpretable representation. In contrast to PCA, which identifies latent directions of maximal variance without preserving semantic meaning (\cite{jolliffe2002principal}), the canonical segmentation maintains a direct correspondence between groups of features and interpretable functional subsystems. This interpretability is not a secondary advantage but a central property in predictive maintenance applications, where understanding the operational origin of predictive signals is often as important as achieving competitive predictive accuracy (\cite{carvalho_systematic_2019}).

From a broader methodological perspective, these findings suggest that semantic segmentation can be interpreted as an information-preserving structural filter. Rather than merely reducing dimensionality, it reorganizes the monitored feature space into dynamically coherent functional groups, preserving the dominant predictive content while isolating variables with weak or non-complementary contributions. This interpretation extends the role of feature engineering beyond performance optimization (\cite{guyon2003introduction}): semantic segmentation becomes a mechanism for revealing the functional organization of the monitored system in a form that is both analytically tractable and operationally interpretable.

At the same time, some limitations must be acknowledged. First, the empirical validation has been performed on a single monitored system and on a limited number of fault events. Although the consistency of the results across validation schemes supports the robustness of the conclusions, broader validation across different infrastructures and failure scenarios is necessary to establish generality. Second, the segmentation rules are defined through domain-informed criteria rather than learned automatically. While this improves interpretability, it may limit portability to systems where semantic relationships among variables are less explicit. Third, the interpretation of predictive dominance is based on empirical risk and conditional dependence, which provide practical evidence of task relevance but do not establish a formal information-theoretic decomposition. Extending the framework toward explicit information measures could provide a stronger theoretical characterization of the relationship between canonical and residual subspaces.

Despite these limitations, the empirical evidence consistently supports the central conclusion of this work: semantic feature segmentation produces an interpretable canonical subspace that preserves the dominant predictive information of the monitored system while isolating weakly informative residual dynamics. The canonical segments exhibit internal structural coherence, remain robust after redundancy reduction, and achieve predictive performance comparable to standard full-space and PCA-based representations. These properties argue that semantic segmentation provides not only an effective feature organization strategy but also a meaningful functional decomposition of the monitored system. In predictive maintenance contexts, where interpretability, robustness, and operational relevance are critical, this dual role represents a substantial methodological advantage.

\section{Summary and conclusions}

This study introduced and empirically evaluated a semantic feature segmentation framework for predictive maintenance, with the intention of assessing whether a domain-informed partition of the monitored variables can reveal an interpretable and functionally meaningful organization of the feature space.

The proposed framework partitions the monitored variables into semantically coherent canonical segments and a residual feature space. The canonical segments group variables according to shared operational roles and expected dynamic behavior, while the residual space collects variables that do not belong to the main semantic subsystems and is treated as an auxiliary control layer. This decomposition was evaluated through a multi-stage validation process addressing structural coherence, robustness under redundancy reduction, and predictive relevance.

The primary finding relates to the structural organization of the canonical space. The empirical analysis showed that intra-segment correlations are systematically higher than inter-segment correlations, indicating that variables grouped within the same segment exhibit stronger coordinated dynamics than variables belonging to different segments. This separability remained statistically significant across the whole canonical space and persisted after redundancy pruning, although with reduced magnitude. Moreover, temporal perturbation experiments demonstrated that the observed separability collapses when cross-feature temporal alignment is disrupted, indicating that the segmentation captures coordinated temporal behavior rather than static statistical similarity. These findings support the interpretation of the canonical segments as dynamically coherent functional subsystems.

The second major result concerns predictive relevance. Across all experimental configurations, the canonical space consistently achieved lower predictive risk than the residual space, showing that semantic segmentation concentrates the dominant predictive information within the canonical representation. At the same time, the residual space exhibited limited predictive capacity and did not provide stable complementary information under high-risk conditions. This suggests that the residual variables mainly capture weakly informative or structurally peripheral dynamics under the current decomposition.

A further important finding emerged from the comparison with standard feature representations. The canonical space achieved predictive performance comparable to both the full feature space and PCA-based dimensionality reduction, suggesting that the semantic segmentation preserves the essential predictive content of the original monitored variables. While PCA provides efficient variance-based compression, semantic segmentation holds interpretability by preserving the correspondence between feature groups and functional subsystems. This characteristic in predictive maintenance, where understanding the operational meaning of predictive signals is often essential for diagnosis and decision support.

Taken together, these results suggest that semantic feature segmentation can act as an information-preserving structural filter: it reorganizes the monitored feature space into interpretable functional groups while maintaining the dominant predictive information required for fault anticipation. This dual role provides a methodological advantage by combining predictive effectiveness with structural interpretability.

The study also highlights important limitations. The validation was conducted on a single monitored system and on a limited number of fault events, so the generality of the conclusions should be interpreted cautiously. In addition, the segmentation rules were defined through domain knowledge rather than learned automatically, which improves interpretability but may reduce portability across heterogeneous systems. Finally, the present study provides an empirical operational characterization, while a formal information-theoretic treatment remains an important future direction..

Future work may extend the proposed framework in several directions. Applying the methodology to additional infrastructures and failure modes would help assess the generality of the observed structural patterns. Integrating domain-informed segmentation with data-driven optimization could improve portability while preserving interpretability. Furthermore, the adoption of explicit information-theoretic measures may provide a stronger theoretical characterization of the relationship between canonical and residual feature spaces.

In conclusion, the empirical evidence suggests that semantic feature segmentation provides a practical and interpretable way to organize monitored variables into dynamically coherent subsystems while preserving predictive performance. Within the experimental conditions considered in this work, the canonical feature space retained the dominant predictive information of the monitored system, whereas the residual space contributed only marginally. These findings indicate that semantic segmentation can support both predictive modeling and structural interpretation, offering a promising methodological direction for interpretable predictive maintenance in complex monitored systems.

\section*{Acknowledgements}
This work is supported by ICSC – Centro Nazionale di Ricerca in
High Performance Computing, Big Data and Quantum Computing,
funded by the European Union – NextGenerationEU.

%% The Appendices part is started with the command \appendix;
%% appendix sections are then done as normal sections
\appendix

\section{Extended Results and Detailed Statistical Analysis}
%% \label{}

%% If you have bibdatabase file and want bibtex to generate the
%% bibitems, please use
%%
%%\clearpage
\bibliographystyle{elsarticle-harv} 
\bibliography{bib_SemSegFea}

%% else use the following coding to input the bibitems directly in the
%% TeX file.

%%\begin{thebibliography}{00}

%% \bibitem[Author(year)]{label}
%% For example:

%% \bibitem[Aladro et al.(2015)]{Aladro15} Aladro, R., Martín, S., Riquelme, D., et al. 2015, \aas, 579, A101

%%\end{thebibliography}
\end{document}